\documentclass{article} 
\usepackage{nips15submit_e,times}
\usepackage{hyperref}
\usepackage{url}
\usepackage[cmex10]{amsmath} 
\usepackage{amssymb}  
\usepackage{subcaption}
\usepackage{caption}
\usepackage{graphicx}
\usepackage{algorithm}
\usepackage{algpseudocode}

\title{Variable-Length Hashing}

\author{
Honghai Yu \\
Data Analytics Department\\
Institute for Infocomm Research\\
Singapore 138632 \\
\texttt{yuhh@i2r.a-star.edu.sg} \\
\And
Pierre Moulin \\
Electrical and Computer Engineering \\
University of Illinois at Urbana-Champaign \\
Urbana, IL 61801\\
\texttt{moulin@ifp.uiuc.edu} 
\And
Hong Wei Ng \\
Advanced Digital Science Center\\
Singapore\\
\texttt{hwng@adsc.com.sg} \\
\And
Xiaoli Li \\
Data Analytics Department\\
Institute for Infocomm Research\\
Singapore 138632 \\
\texttt{xlli@i2r.a-star.edu.sg} \\
}

\nipsfinalcopy 

\begin{document}

\maketitle

\begin{abstract}
Hashing has emerged as a popular technique for large-scale similarity search. Most learning-based hashing methods generate compact yet correlated hash codes. However, this redundancy is storage-inefficient. Hence we propose a lossless variable-length hashing (VLH) method that is both storage- and search-efficient. Storage efficiency is achieved by converting the fixed-length hash code into a variable-length code. Search efficiency is obtained by using a multiple hash table structure. With VLH, we are able to deliberately add redundancy into hash codes to improve retrieval performance with little sacrifice in storage efficiency or search complexity. In particular, we propose a block K-means hashing (B-KMH) method to obtain significantly improved retrieval performance with no increase in storage and marginal increase in computational cost.
\end{abstract}

\section{Introduction}
Retrieval of similar objects is a key component in many applications such as large-scale visual search. As databases grow larger, learning compact representations for  efficient storage and fast search becomes increasingly important. These representations should preserve similarity, i.e., similar objects should have similar representations. Hashing algorithms, which encode objects into compact binary codes to preserve similarity, are particularly suitable for addressing these challenges.


The last several years have witnessed an accelerated growth in hashing methods. One common theme among these methods is to learn compact codes because they are storage and search efficient. In their pioneering work on spectral hashing (SH) \cite{Weiss08}, Weiss et al argued that independent hash bits lead to the most compact codes, and thus independence is a desired property for hashing. Some other hashing methods such as \cite{Indyk98,Shakhnarovich03,Raginsky09,WangCVPR10,Liu12,Liu13,YuICASSP15}, explicitly or implicitly aim at generating independent bits. However, it is often difficult to generate equally good independent bits. Under the unsupervised setting, it has been observed that data distributions are generally concentrated in a few high-variance projections \cite{Wang10,Gong11}, and performance deteriorates rapidly as hash code length increases \cite{liu14}. Under the supervised setting, it has also been noticed that the number of high signal-to-noise ratio (SNR) projections is limited, and bits generated from subsequent uncorrelated low-SNR projections may deteriorate performance \cite{Yu15}. Therefore, most  hashing methods generate correlated bits. Some learn hash functions sequentially in boosting frameworks \cite{Shakhnarovich05,Wang10,Lin10}, some learn orthogonal transformations on top of PCA projections to balance variances among different projection directions \cite{Gong11,Kong12}, some learn multiple bits from each projection \cite{Liu11,Yu15}, and many others learn hash functions jointly without the independence constraint \cite{KulisNIPS09,Weiss12,Norouzi12}.

One drawback of correlated hash codes is the storage cost caused by the redundancy in the codes. Surprisingly, this drawback has never been addressed, even though one of the main purposes of hashing is to find storage-efficient representations. Theoretically, this redundancy could be eliminated by entropy coding \cite{Cover:1991}, where more frequent patterns are coded with fewer bits and less frequent patterns are coded with more bits. Practically, entropy coding faces two major challenges: (1) as the number of codewords is exponential in the input sequence length $B$, it is infeasible to estimate the underlying distribution when $B$ is large; (2) as entropy coding produces variable-length codes, it is not clear how to find nearest neighbors of a query without first decoding every database item to the original fixed-length hash codes, which will increase search complexity tremendously. Perhaps due to these two challenges, all existing hashing methods require data points to be hashed into the same number of bits, which leave little room for redundancy reduction. The first contribution of this paper is to propose a two-stage procedure, termed variable-length hashing (VLH), that is not only capable of reducing redundancy in hash codes to save storage but also is search efficient. The first stage is a lossless variable-length encoder that contains multiple sub-encoders, each of which has moderate complexity. The second stage is a multiple hash table data structure that combines the variable-length codes from stage 1 and the multi-index hashing algorithm \cite{Norouzi12MultiIndex} to achieve search efficiency.

On the other hand, deliberately adding redundancy into a system boosts performance in many applications. For instance, channel coding  uses extra bits to improve the robustness to noise in digital communication systems \cite{Cover:1991}, sparse coding uses overcomplete dictionary to represent images for denoising, compression, and inpainting \cite{Aharon06}, and content identification for time-varying sequences uses overlapping frames to overcome the desynchronization problem \cite{Haitsma02}. The second contribution of this paper is to demonstrate the effectiveness of adding redundancy in hashing, and shed some light on this new design paradigm for hashing. Specifically, we propose a block K-means hashing (B-KMH) method, in the spirit of block codes in channel coding, to represent each K-means codeword with more than the necessary number of bits so that the Hamming distance between hash codes can better approximate the Euclidean distance between their corresponding codewords. B-KMH is an extension to the state-of-the-art K-means hashing (KMH) \cite{He13}. On two large datasets containing one million points each,  we demonstrate B-KMH's superior approximate nearest neighbor (ANN) search performance over KMH and many other well-known hashing methods. Moreover, the added redundancy can be removed from storage with only marginal increase in search complexity using VLH.

\section{Lossless Compression by Variable-Length Hashing}
In this section, we first  propose a variable-length encoding scheme that encodes fixed-length hash codes into variable lengths thus reducing the average code length. Moreover, we show that the variable-length codes can be seamlessly combined with multi-index hashing \cite{Norouzi12MultiIndex} to efficiently find nearest neighbors. 
\subsection{Variable-Length Encoding} 
\label{Sec:VLE}
Let us consider a 64-bit hash code, for example produced by ITQ \cite{Gong11}. The expected code length in this fixed-length setting is $L=64$ bits. We know that bits are correlated as they are generated from correlated projections. An entropy coder, such as the Huffman coder, could achieve an expected code length $L < 64$ bits without any information loss. However, to use entropy coding, one needs to estimate the probabilities for $K =2^{64}$ symbols, which would require many times $K$ examples. Moreover, it is impossible to store the codebook consisting of $K$ codewords.

Inspired by the product quantizer \cite{Sabin84,Jegou11} which allows us to choose the number of components to be quantized jointly, we partition the $B$-bit hash code $\textbf{f} \in \{0,1\}^B$ into $M$ distinct binary substrings $\textbf{f} = \{\textbf{f}^{(1)},\ldots,\textbf{f}^{(M)}\}$. For convenience, we assume $M$ divides $B$ and each substring consists of $b=B/M$ bits. This limitation can be easily circumvented. The substrings are compressed separately using $M$ distinct encoders. For each substring, the number of symbols is only $2^{b}$. For instance, each substring has only 256 unique symbols when a 64-bit code is partitioned into 8 substrings, and can be learned accurately using one million examples. Moreover, the total number of codewords is only $M$ times $2^{b}$. Note that in the extreme case where $M=B$, the bits are all encoded separately and the variable-length procedure reduces to the original fixed-length hashing procedure.

As will be shown in the next section, variable-length codewords are stored in different hash tables for retrieval with one hash table corresponding to one encoder. This multiple hash table data structure provides great flexibility in designing encoders because codewords are naturally separated. Therefore, we can drop the {\em prefix code} constraint, and only require codewords to be different (such codes are called {\em nonsingular} \cite{Cover:1991}) to ensure unique decodability. Here, we propose the following encoding procedure for each substring:
\begin{enumerate}
\item Estimate the probability $p_i$ for each of the $2^b$ symbols. Assign a  small probability $\epsilon$ to symbols not appearing in the database.
\item  Rearrange symbols in decreasing order of $\{p_i\}$. Starting from one bit, assign one more bit to the next most probable symbol if all shorter bit strings have been assigned. To ensure each codeword can be converted into a unique integer \footnote{Thus, we can simply use the corresponding integer as the index to the decoding table, which makes decoding super fast.}, bit strings longer than one bit and with the most significant bit (MSB) equal to ``0" will not be used as codewords. Therefore, the codewords for the first five most probable symbols are ``0", ``1", ``10", ``11", and ``100".
\end{enumerate}
One advantage of the proposed encoder is that the maximum codeword length does not exceed the substring length $b$, which gives greater control over the size of the decoding table. 
\begin{figure*}[h] 
        \centering
				$
        \begin{array}{cc}

        \begin{subfigure}[b]{0.45\textwidth}
                \centering
                \includegraphics[width=\textwidth]{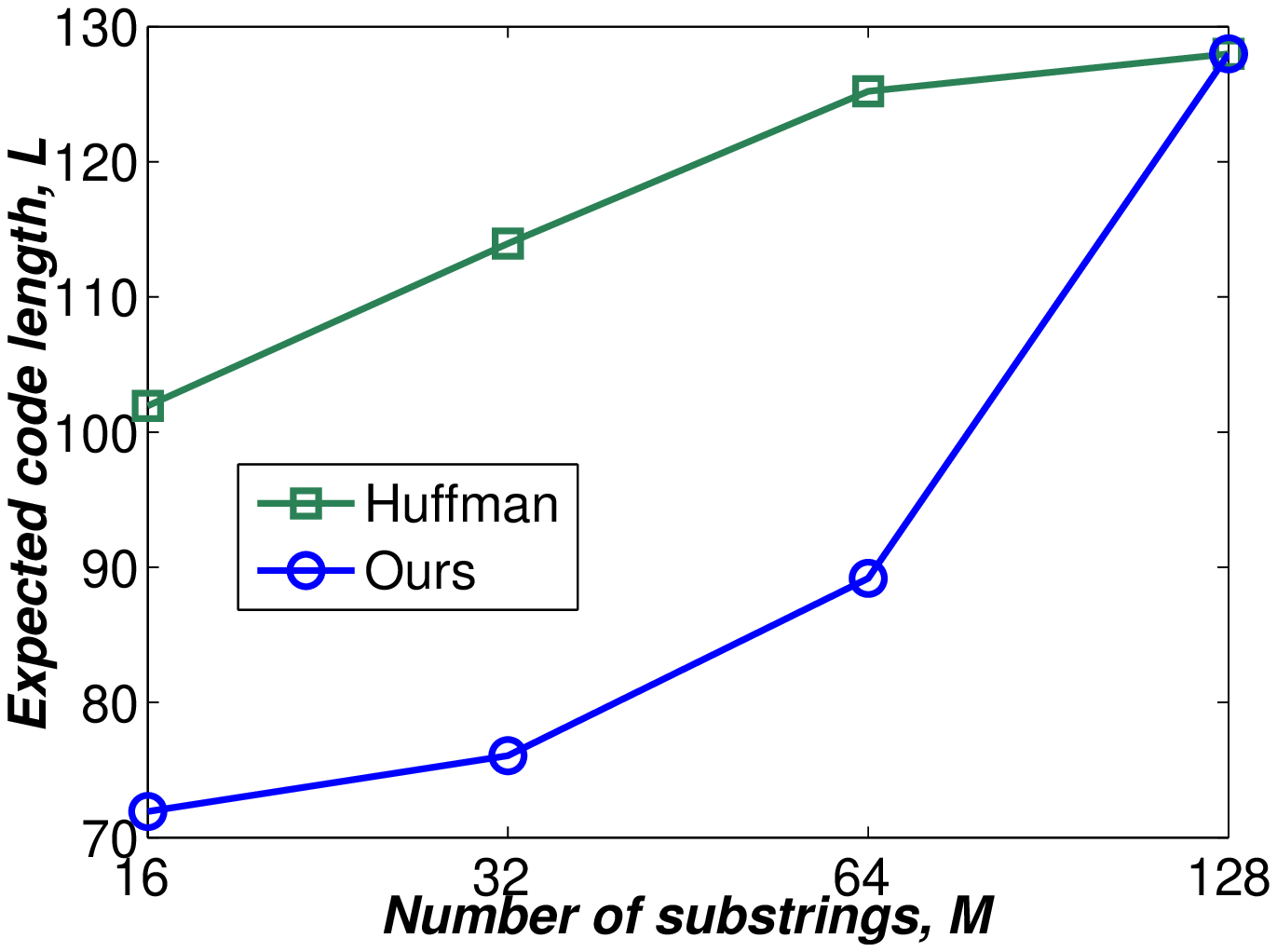}
               \caption{\small{Expected code length for different $M$.} }
               \label{fig:Ave_L}
        \end{subfigure} 
				~\quad
				\begin{subfigure}[b]{0.45\textwidth}
                \centering
                \includegraphics[width=\textwidth]{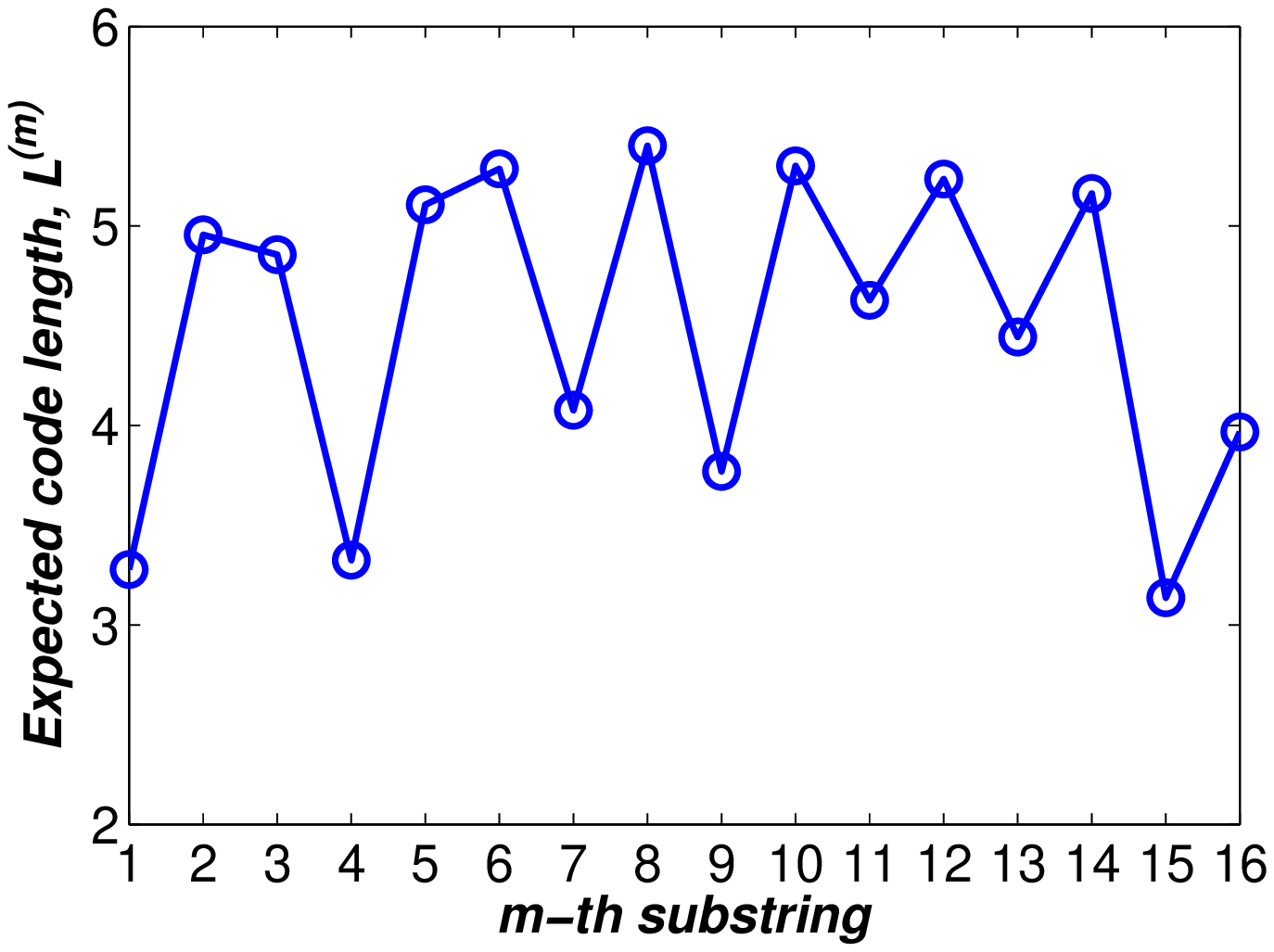} 
                \caption{\small{$L^{(m)}$ for $m=1,\ldots,16$.}}
                \label{fig:Lm}
        \end{subfigure}%
        \end{array}$ 
        \caption{\small{Variable-length encoding of 128-bit ITQ hash codes on the SIFT1M dataset. $M=128$ corresponds to the fixed-length codes.}}
        \label{fig:1M_SIFT_L}
\end{figure*}

For the $m$-th substring, each symbol will be assigned a codeword with length $l_i^{(m)}$, $1 \leq i \leq 2^{b}$, resulting an expected code length of 
\begin{equation}
L^{(m)} = \sum_{i=1}^{2^{b}} p^{(m)}_i l_i^{(m)}.
\end{equation}
Thus, the overall expected code length is $L = \sum_{m=1}^M L^{(m)}$. 

As an example, let us examine how much compression can be done to the hash codes generated by ITQ on the SIFT1M dataset \cite{Jegou11}, which contains one million 128-D SIFT descriptors \cite{Lowe04}. From each data point in the SIFT1M dataset, ITQ extracts a 128-bit hash code. As shown in Fig.~\ref{fig:Ave_L}, our encoder has a much higher compression ratio than the Huffman encoder, and expected code lengths decrease  when longer substrings are jointly compressed (corresponding to a smaller $M$). Fig.~\ref{fig:Lm} shows the substring expected code length $L^{(m)}$ for $1 \leq m \leq 16$ in our encoder.  It is clear that different substrings contain different degrees of redundancy. The significant improvement over the optimal prefix code, i.e., the Huffman code, is due to our multiple hash table data structure, which enables us to us nonsingular codes, a superset of prefix codes. Note that Fig.~\ref{fig:1M_SIFT_L} shows the theoretical amount of compression each encoder can achieve. In practice, the amount of saving is subject to the smallest unit of representation in computer architectures. 

\subsection{Multi-Index Hashing}
\label{sec:MIH}
Multi-index hashing (MIH) \cite{Norouzi12MultiIndex} is an efficient algorithm for exact nearest neighbor search on hash codes. It has a provably sub-linear search complexity for uniformly distributed codes. Practice has shown it to be more than 100 times faster than a linear scan baseline on many large-scale datasets \cite{Norouzi12MultiIndex}. However in its formulation, MIH assumes fixed-length hash codes, and thus considers the storage cost of the hash codes as irreducible. The structure of MIH is compatible with our variable-length encoding. Both rely on partitioning hash codes into binary substrings. In MIH, hash codes from the database are indexed $M$ times into $M$ different hash tables, based on $M$ disjoint binary substrings. With variable-length encoding, we only store the variable-length codes in the hash table buckets, while the fixed-length binary substrings are used as keys to the hash tables. From the previous section, it is clear that we can reduce storage cost with variable-length codes. The rest of this section shows how to combine variable-length encoding with MIH to achieve fast search.

\begin{figure*}[t] 
        \centering
				$
        \begin{array}{c}
        \begin{subfigure}[b]{0.59\textwidth}
                \centering
                \includegraphics[width=\textwidth]{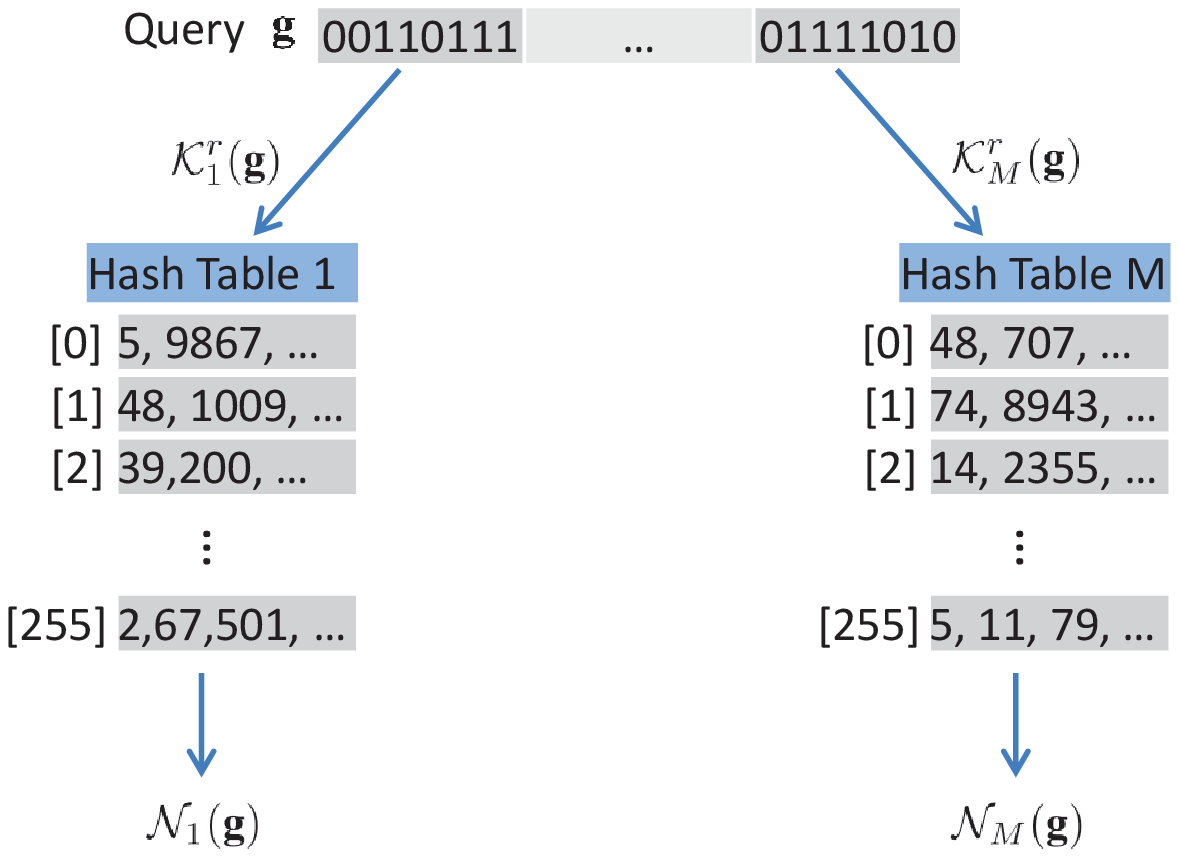}
               \caption{\small{Hash table loolup.} }
               \label{fig:lookup}
        \end{subfigure} 
				\\
				\begin{subfigure}[b]{0.9\textwidth}
                \centering
                \includegraphics[width=\textwidth]{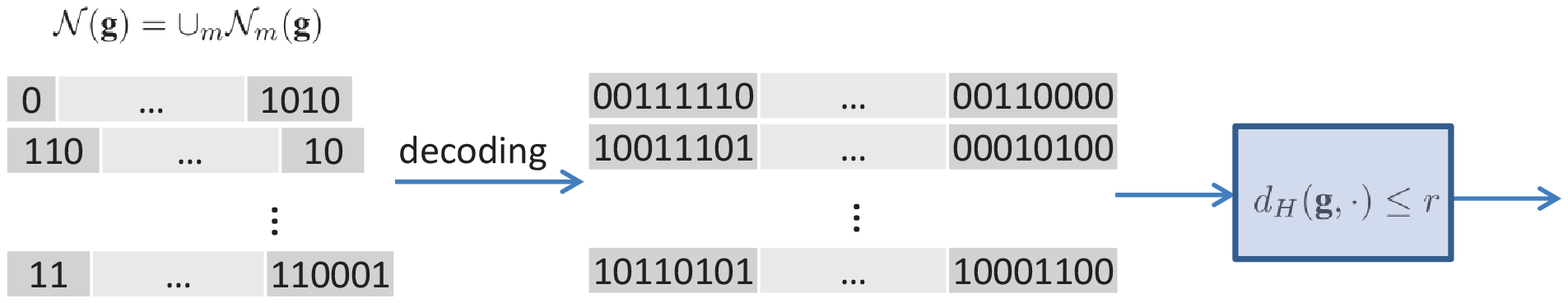} 
                \caption{\small{Candidate test.}}
                \label{fig:candidate}
        \end{subfigure}%
        \end{array}$ 
        \caption{\small{Multi-index hashing with variable-length codes.}}
        \label{fig:search}
\end{figure*}

The key idea of MIH rests on the following proposition: When two hash codes $\textbf{f}$ and $\textbf{g}$ differ by $r$ bits or less, then, in at least one of their $M$ substrings they must differ by at most $\left \lfloor r/M \right \rfloor $ bits. This follows straightforwardly from the Pigeonhole Principle \cite{Norouzi12MultiIndex}. Given a query $\textbf{g} = \{\textbf{g}^{(1)},\ldots,\textbf{g}^{(M)}\}$, MIH finds all its $r$-neighbors (that is all samples $\textbf{f}$ such that $d_H(\textbf{g}, \textbf{f}) <= r$) by first retrieving samples in the database that might be its $r$-neighbors (which we refer to as {\em candidates}) using table lookups followed by testing them exhaustively to check if they indeed fall within a radius $r$ from $\textbf{g}$. This procedure is described in greater detail below and is depicted graphically in Fig.~\ref{fig:search}.

\textbf{Hash table lookup.} For a query $\textbf{g} = \{\textbf{g}^{(1)},\ldots,\textbf{g}^{(M)}\}$, we search the $m$-th substring hash table for entries that are within Hamming distance $\left \lfloor r/M \right \rfloor$ of $\textbf{g}^{(m)}$, i.e., the $m$-th set of search keys $\mathcal{K}_m^r(\textbf{g})$ is given by
\begin{equation}
\mathcal{K}_m^r(\textbf{g}) = \{\textbf{g}' \in \{0,1\}^{b} | d_H(\textbf{g}^{(m)},\textbf{g}') <= \left \lfloor r/M \right \rfloor\},
\end{equation}
where the integer value of each search key in $\mathcal{K}_m^r(\textbf{g})$ gives us the hash table index (inside square brackets in Fig.~\ref{fig:lookup}) to look for matching substrings. Thus, for each hash table, we retrieve a set of candidates $\mathcal{N}_m(\textbf{g})$. Based on the proposition above, the union of the candidate sets $\mathcal{N}(\textbf{g}) = \cup_m \mathcal{N}_m(\textbf{g})$ is a superset of the $r$-neighbors of $\textbf{g}$. In this step, the run-time cost is the same for fixed-length and variable-length codes, while variable-length codes has a lower storage cost.

\textbf{Candidate test.} To find the true $r$-neighbors of $\textbf{g}$, we need to compute the Hamming distance between $\textbf{g}$ and every hash code in $\mathcal{N}(\textbf{g})$. However before doing so, variable-length codes need to be decoded into fixed-length codes. To decode each codeword, we need $M$ table lookups. As decoding tables are typically small, e.g., for a 16-bit substring the decoding table is only 128 Kbytes, they can be load into cache memory for fast decoding. 

\begin{figure*}[h] 
        \centering
				$
        \begin{array}{cc}

        \begin{subfigure}[b]{0.45\textwidth}
                \centering
                \includegraphics[width=\textwidth]{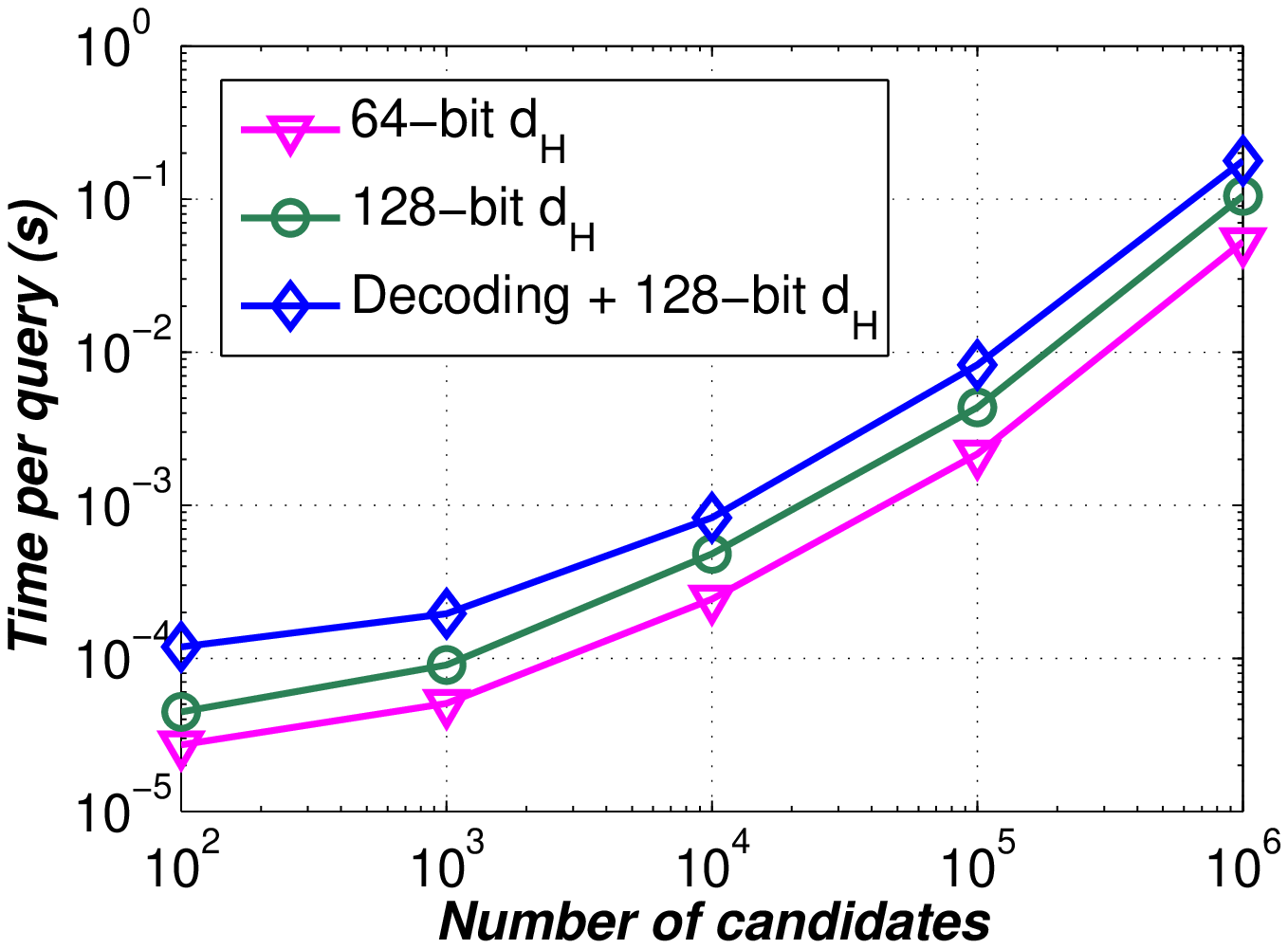}
                \caption{\small{Candidate test cost.}}
                \label{fig:time_per_query}
        \end{subfigure} 
        ~\quad
        \begin{subfigure}[b]{0.45\textwidth}
                \centering
                
                \includegraphics[width=\textwidth]{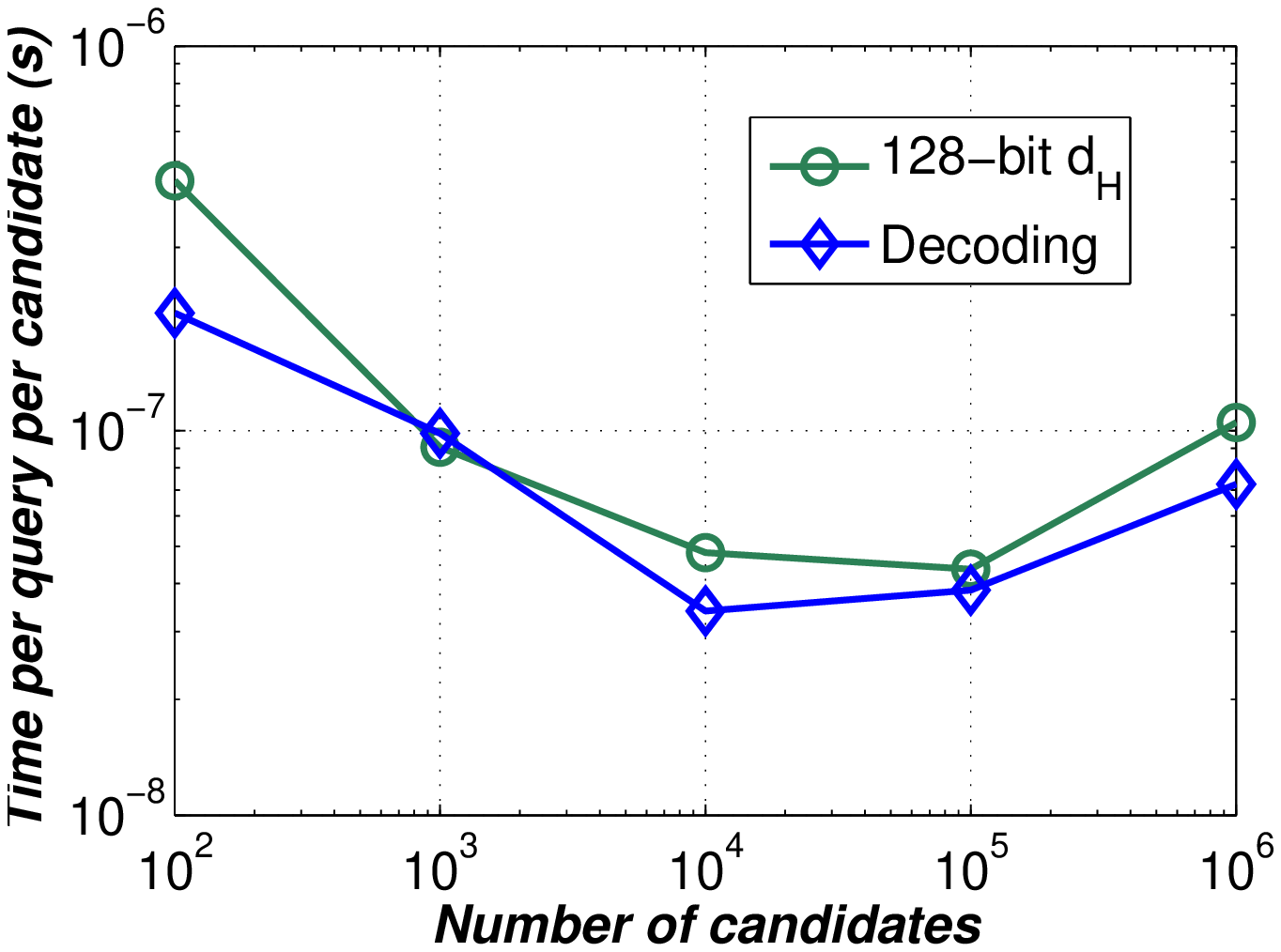}
                \caption{\small{Decoding vs Hamming distance computation.}}
                \label{fig:time_per_query_per_candidate}
        \end{subfigure} 
        \end{array}$ 
        \caption{\small{Run-time comparisons for candidate test between fixed-length and variable-length hash codes at various candidate set size. We use $(M=8,b=16)$ for encoding 128-bit codes.}}
        \label{fig:SIFT1M_runTime}
\end{figure*}

To compare the run-time of decoding and Hamming distance computation $d_H$, we run 1,000 simulations using MATLAB on an office desktop with Intel Core i7 3.40 GHz and 8GB RAM. We use substring of length $b=16$ for encoding 128-bit codes, resulting in $M=8$ small decoding tables. In Fig.~\ref{fig:time_per_query}, we show the average run-time per query with up to one million candidates. Decoding one million candidate codewords followed by one million Hamming distance computation costs merely 0.17 sec for 128-bit hash codes.  However, as the number of candidates $|\mathcal{N}(\textbf{g})|$ is typically much smaller than the database size, we should expect the candidate test cost be much smaller than 0.17 sec for most databases \cite{Norouzi12MultiIndex}. In Fig.~\ref{fig:time_per_query_per_candidate}, we see that the average decoding run-time is actually smaller than Hamming distance computation, so decoding will not be the bottleneck in real-time search applications. Note that MATLAB is not optimized for speed, we expect the actual run-times to be much faster than those shown in Fig.~\ref{fig:SIFT1M_runTime}.

\section{Improving Retrieval Performance by Adding Redundancy}
Knowing redundancy could be reduced by VLH with little increase in search complexity, we would like to examine whether deliberately adding redundancy into hash codes would improve retrieval performance. We test this idea on the recent K-means hashing (KMH) \cite{He13} method, which has shown excellent ANN search performance, and propose a block K-means hashing (B-KMH) method to represent each K-means codeword with more than the necessary number of bits so that the Hamming distance between binary representations can better approximate the Euclidean distance between their corresponding codewords. 

\subsection{K-means Hashing}
Recently, vector quantization has shown excellent ANN search performance \cite{Jegou11}. It approximates the Euclidean distance between two vectors $\textbf{x}$ and $\textbf{y}$ by the Euclidean distance between their codewords:
\begin{equation}
d(\textbf{x},\textbf{y}) \approx  d(\textbf{c}_{i(\textbf{x})},\textbf{c}_{i(\textbf{y})}),
\end{equation}
where $i(\textbf{x})$ denotes the index of the cell containing $\textbf{x} \in \mathbb{R}^d$, and $\textbf{c}_{i(\textbf{x})} \in \mathbb{R}^d$ is the codeword. However, vector quantization is considerably slower than Hashing-based methods \cite{He13}. To take advantage of fast Hamming distance computation, k-means hashing (KMH) approximates the pairwise codewords distance by the Hamming distance between the codewords' binary representations (hash codes): 
\begin{equation}
\label{eqn:dist_appr}
d(\textbf{c}_{i(\textbf{x})},\textbf{c}_{i(\textbf{y})}) \approx d_s(I_{i(\textbf{x})},I_{i(\textbf{y})}),
\end{equation} 
where $I_i \in \{0,1\}^b$ is the $b$-bit representation of $\textbf{c}_i$, and $d_s$ is defined as a rescaled Hamming distance between any two binary representations $I_i$ and $I_j$:
\begin{equation}
d_s(I_i,I_j) \triangleq s \cdot d_H^{\frac{1}{2}}(I_i,I_j),
\end{equation}
where $s$ is a constant scale and $d_H$ denotes the Hamming distance. The square root is necessary because it enables to generalize this approximation to product space. The use of $s$ is because the Euclidean distance can be in an arbitrary range, while the Hamming distance $d_H$ is constrained in $[0,b]$ given $b$ bits.

Given a training dataset $\mathcal{X} \in \mathbb{R}^{n \times d}$ consisting of $n$ $d$-dimensional vectors, KMH learns $k = 2^b$ codewords corresponding to $k$ cells.  Note that $\{I_i\}$ are predetermined and remain the same throughout the training process. KMH considers two error terms: the average {\em quantization error} $E_{\text{quan}}$ 
\begin{equation}
E_{\text{quan}} = \frac{1}{n} \sum_{\textbf{x} \in \mathcal{X}} ||\textbf{x} - \textbf{c}_{i(\textbf{x})}||^2,
\end{equation}
and the {\em affinity error} $E_{\text{aff}}$, which is the average error due to the distance approximation in (\ref{eqn:dist_appr}), 
\begin{equation}
E_{\text{aff}} = \sum_{i=1}^k \sum_{j=1}^k w_{ij} (d(\textbf{c}_i,\textbf{c}_j) - d_s(I_i,I_j))^2,
\end{equation}
where $w_{ij} = n_in_j/n^2$, and $n_i$ and $n_j$ are the number of samples having index $i$ and $j$ respectively. The overall cost function of KMH is
\begin{equation}
\label{kmeansObj}
E = E_{\text{quan}} + \lambda E_{\text{aff}},
\end{equation}
where $\lambda$ is a fixed weight. Minimizing (\ref{kmeansObj}) takes an alternating fashion:
\begin{itemize}
\item
{\em Assignment step: fix $\{\textbf{c}_i\}$ and optimize $i(\textbf{x})$}. Each sample $\textbf{x}$ is assigned to its nearest codeword in the  codebook $\{\textbf{c}_i\}$. This is the same as the K-means algorithm.
\item
{\em Update step: fix  $i(\textbf{x})$ and optimize $\{\textbf{c}_i\}$}. Each codeword $\textbf{c}_j$ is sequentially optimized with other $\{\textbf{c}_i\}_{i\neq j}$ fixed.
\end{itemize}

Similar to Product Quantization \cite{Jegou11}, KMH is easily generalized to a Cartesian product of subspaces, where (\ref{kmeansObj}) is independently minimized in each subspace \cite{He13}. To keep model complexity manageable, each subspace will be assigned a small number of bits, often not exceeding $b=8$ corresponding to at most 256 codewords.

\subsection{Block K-means Hashing}
\begin{figure*}[th] 
        \centering
				$
        \begin{array}{ccc}

        \begin{subfigure}[b]{0.32\textwidth}
                \centering
                \includegraphics[width=\textwidth]{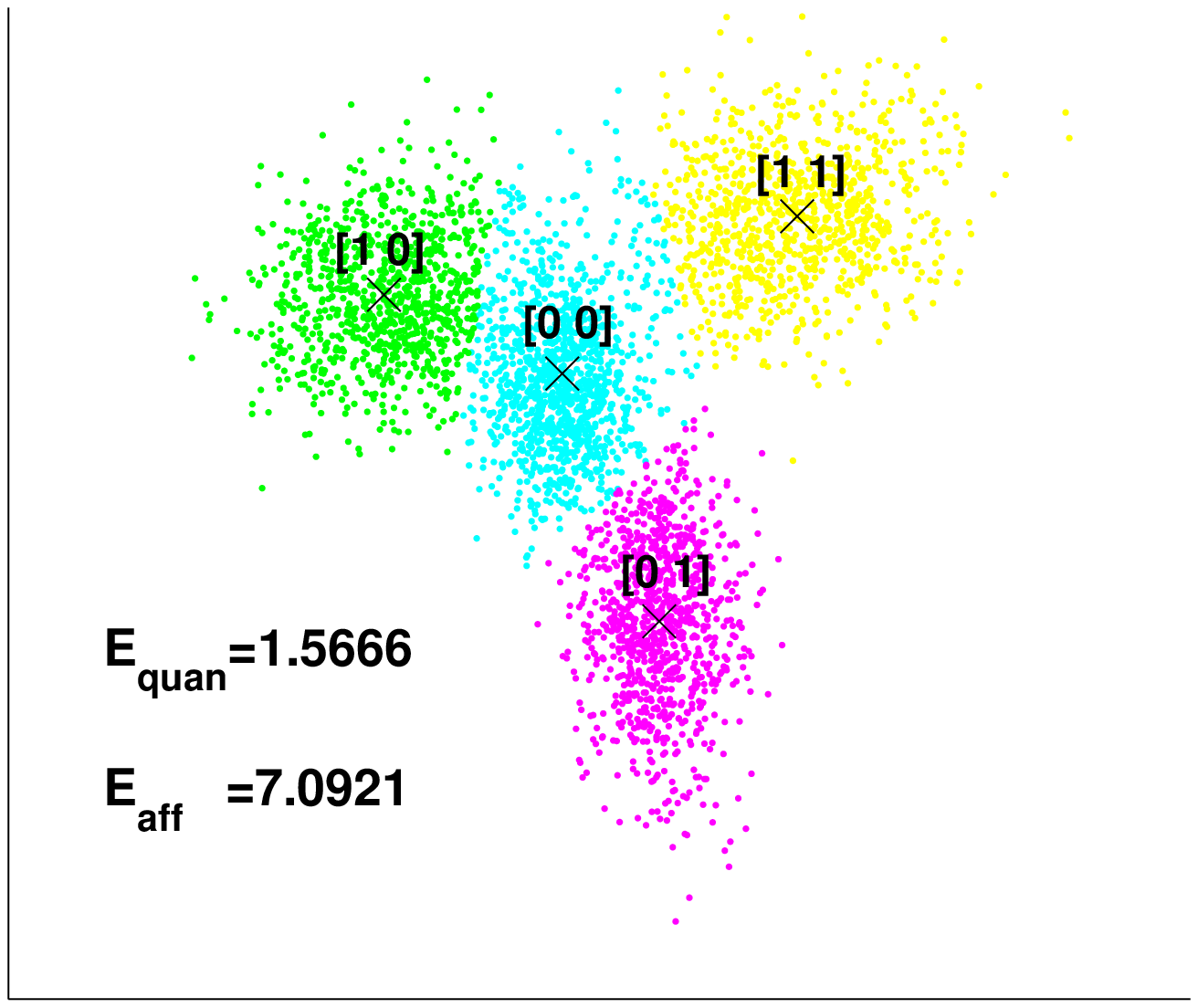}
               \caption{\small{Naive Two-Step} }
               \label{fig:NTS}
        \end{subfigure} 
				~
				\begin{subfigure}[b]{0.32\textwidth}
                \centering
                \includegraphics[width=\textwidth]{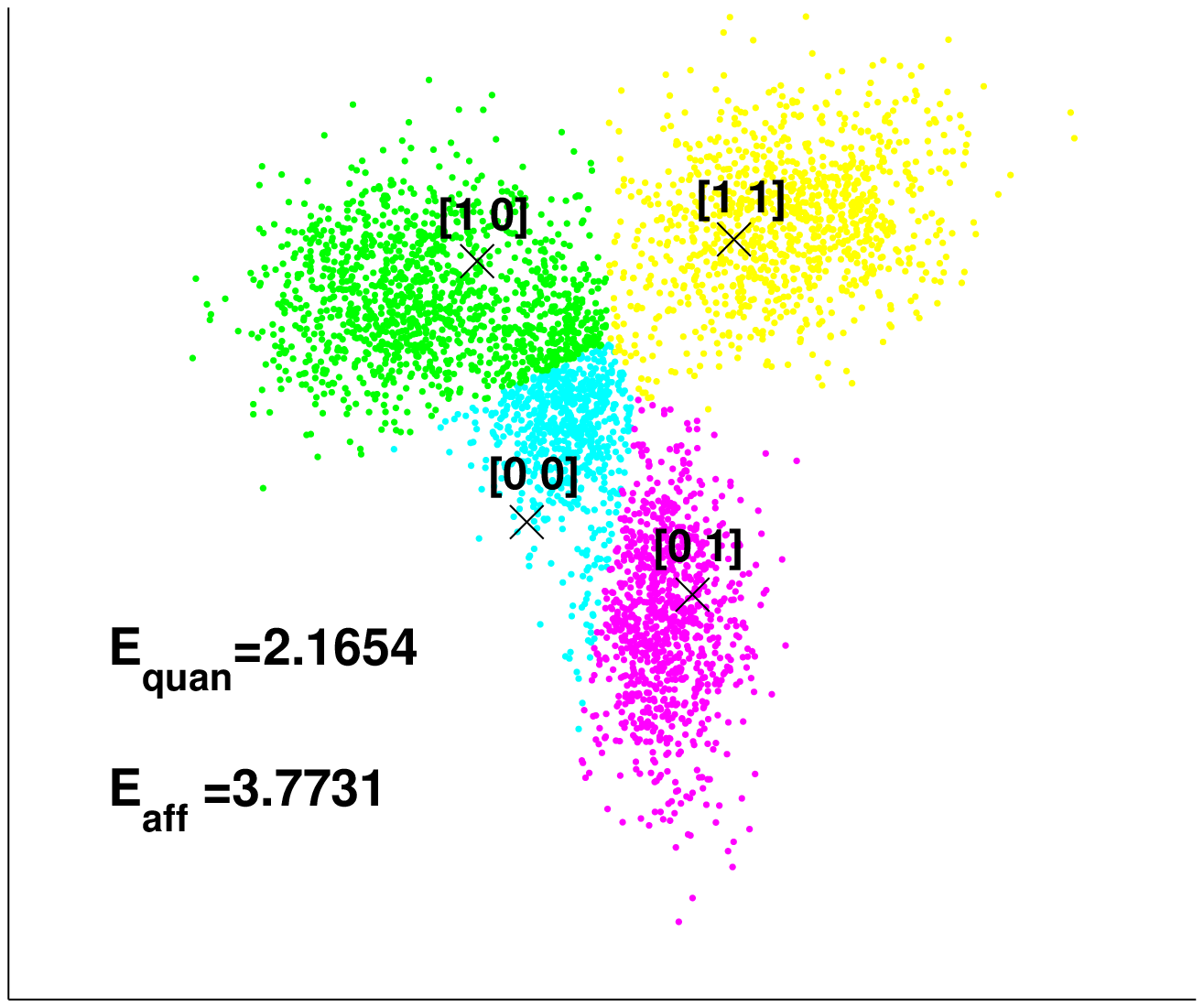} 
                \caption{\small{KMH.}}
                \label{fig:KMH}
        \end{subfigure}%
        ~
         \begin{subfigure}[b]{0.32\textwidth}
                \centering
                \includegraphics[width=\textwidth]{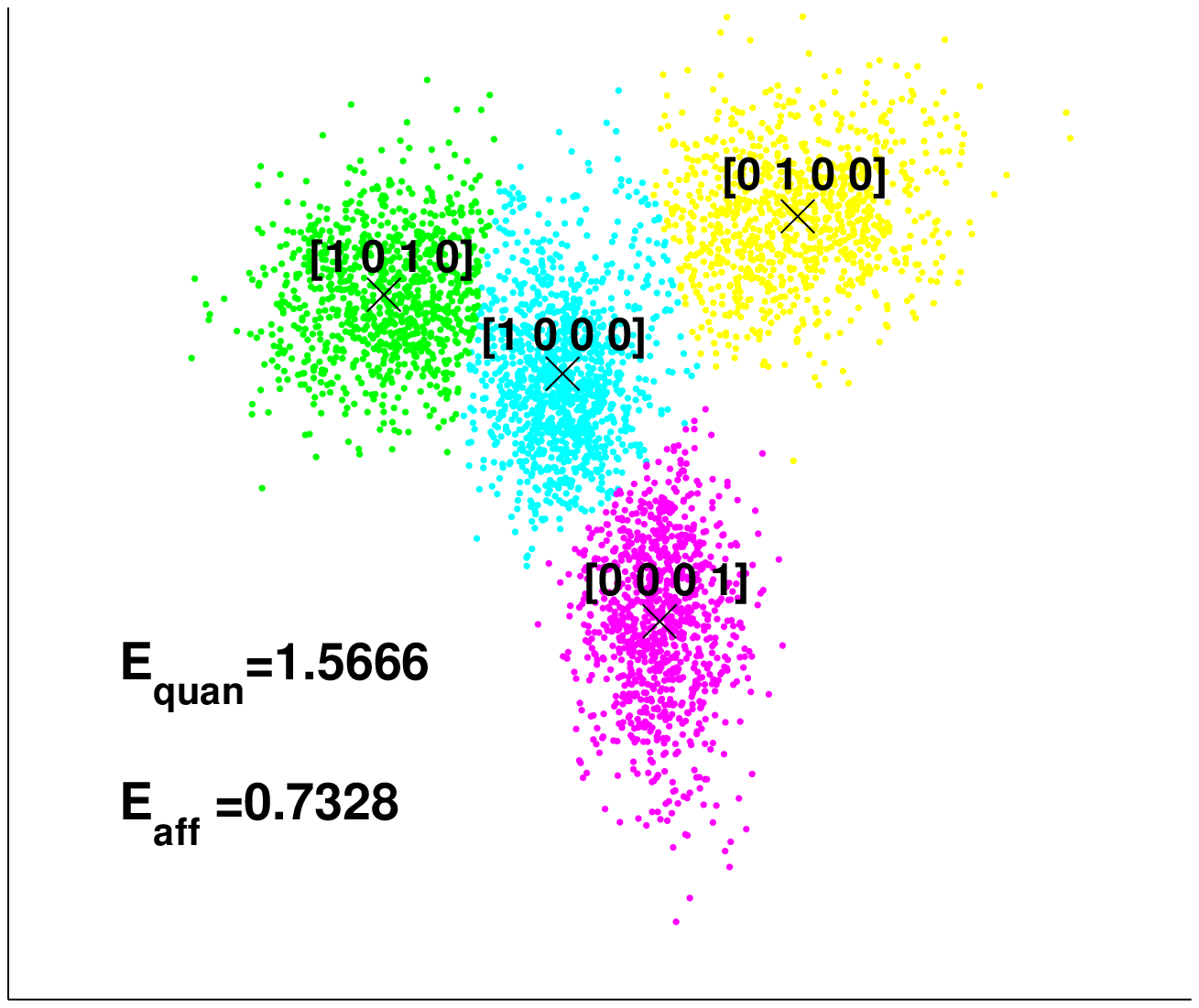}
               \caption{\small{B-KMH.} }
               \label{fig:BKMH}
        \end{subfigure} 
        \end{array}$ 
        \caption{\small{Illustration of the proposed B-KMH (best viewed in color). B-KMH achieves both the smallest quantization error and affinity error.}}
        \label{fig:simulation}
\end{figure*}

\begin{algorithm}[h]
\caption{Block K-means hashing (B-KMH)}
\label{Alg:BKMH}
\textbf{Input}: Training dataset $\mathcal{X} \in \mathbb{R}^{n \times d}$, number of codewords $k$, and all $\beta$-bit strings $\mathcal{B} = \{0,1\}^\beta$
\begin{algorithmic}[1]
\item Learn $\{\textbf{c}_i\}_{i=1}^k = \arg \min E_{\text{quan}}$ by K-means.
\item Randomly choose $k$ different $\beta$-bit strings $\{I_i\}_{i=1}^k$ from $\mathcal{B}$ to represent $\{\textbf{c}_i\}_{i=1}^k$.
\item Obtain $s = \arg \min E_{\text{aff}}$ by solving a quadratic function.
\While {termination conditions are not met}
\For{i=1:k}
\State $I_i=\arg \min_{I_i \in \mathcal{B}} E_{\text{aff}}$
\EndFor
\EndWhile
\end{algorithmic}
\textbf{Output}: $\{\textbf{c}_i\}$ and $\{I_i\}$
\end{algorithm}
In KMH, $b$ bits are used to represent $k=2^b$ codewords. However, this representation can be very restrictive as the number of distinct Hamming distances between any two $b$-bit strings is only $b+1$. To make $E_{\text{aff}}$ small, KMH will need to force the codewords to respect the simple geometry of their binary representations, resulting in a large $E_{\textbf{quan}}$. Therefore, we propose to use more than $b$ bits to represent $k$ codewords. 

In B-KMH, we represent each of the $k$ codewords with $\beta >\log_2 k$ bits. The goal of B-KMH is to search for the optimal set of $k$ representations among the $2^\beta$ binary strings $\mathcal{B} = \{0,1\}^\beta$ to minimize $E_{\text{aff}}$. Moreover, the codebook $\{\textbf{c}_i\}$ is learned by minimising $E_{\text{quan}}$, which is solved by the K-means algorithm, and stays unchanged during the training. Therefore, the differences between KMH and B-KMH are: (1) binary representations are fixed in KMH, while they are learned in B-KMH; (2) codewords are learned in KMH, while they are fixed by the K-means algorithm in B-KMH; (3) the objective function of KMH is $E = E_{\text{quan}} + \lambda E_{\text{aff}}$, while B-KMH uses only $E_{\text{aff}}$.

Exhaustively search all possible combinations of $k$ representations from $2^\beta$ candidates is feasible only for very small $k$ and $\beta$. Therefore, B-KMH relies on a greedy search strategy. It randomly choose $k$ elements from $\mathcal{B}$ to form an initial set of representations $\{I_i\}$, and initialize $s$ by minimizing $E_{\text{aff}}$ with respect to $s$. We fix $s$ after initialization. Next, we update each $I_j$ by exhaustively searching $\mathcal{B}$ to minimize $E_{\text{aff}}$ while $I_i, i \neq j$ being fixed. The sequential updates will terminate when there is no change in $\{I_i\}$ or the number of iterations exceeds a predefined threshold. Moreover, we could run B-KMH multiple times with different initializations, and choose the one generates the smallest $E_{\text{aff}}$. A summary of the proposed B-KMH method is presented in Algorithm \ref{Alg:BKMH}.

In KMH \cite{He13}, He et al have also considered a similar strategy that is to use K-means to learn codewords and then assign binary representations to each codeword. However, they only considered using $b=\log_2 k$ bits to represent the codewords, and called this strategy a ``naive two-step" method. Even with exhaustive assignment, their ``naive two-step" method performs much worse than KMH (Fig.~4 in \cite{He13}). We also observe similar behavior in our experiments when setting $\beta = \log_2 k$. However, as $\beta$ exceeds $\log_2 k$, B-KMH outperforms KMH significantly (refer to Fig.~\ref{fig:SIFT1M_beta}). 

In Fig.~\ref{fig:simulation}, we illustrate the differences among the naive two-step, KMH, and B-KMH on a synthetic dataset. Data points are generated by a Gaussian mixture model, and then projected onto the PCA projection directions (KMH uses PCA for initialization.). In Fig.~\ref{fig:NTS}, K-means learned codewords lead to a small quantization error, but fitting $b=2$ bits into $k=4$ codewords is challenging which leads to a large affinity error. For example, the Euclidean distance between codewords of yellow (top right cell) and cyan (middle cell) is much smaller than  that of yellow and magenta (bottom cell), but their Hamming distances are the other way around. In Fig.~\ref{fig:KMH}, KMH learns codewords to respect the geometry of their binary representations, which forms a square in a 2-bit representation. KMH reduces the affinity error but incurs a larger quantization error. If the weight $\lambda = \infty$, the codewords will coincide with their binary representations, resulting in zero affinity error but very large quantization error. On the other hand, B-KMH is able to maintain the small K-means quantization error while reducing the affinity error significantly, as demonstrated in Fig.~\ref{fig:BKMH}. 

Note that the $(\beta - b)$ redundant bits can be removed by VLH, so B-KMH incur no extra storage cost. The additional decoding and Hamming distance with longer codes only apply to the candidate set which is normally much smaller than the database size. As shown in Fig.~\ref{fig:time_per_query}, this increase in search complexity is only marginal in real-time search applications.

\begin{figure*}[ht] 
        \centering
				$
        \begin{array}{cc}

        \begin{subfigure}[b]{0.445\textwidth}
                \centering
                \includegraphics[width=\textwidth]{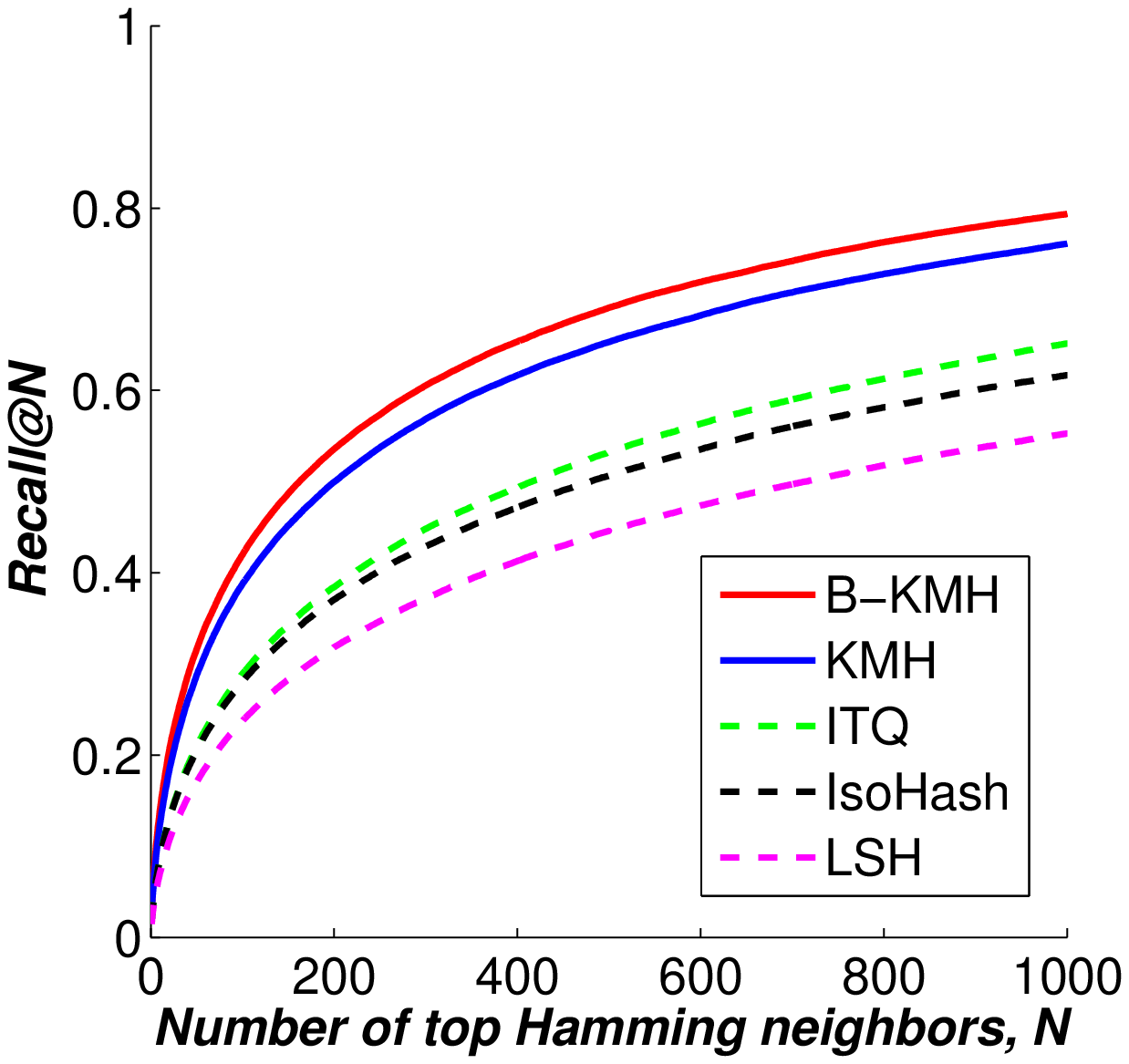}
               \caption{\small{SIFT1M 64-bit.} }
               \label{fig:SIFT1M_64}
        \end{subfigure} 

				\begin{subfigure}[b]{0.445\textwidth}
                \centering
                \includegraphics[width=\textwidth]{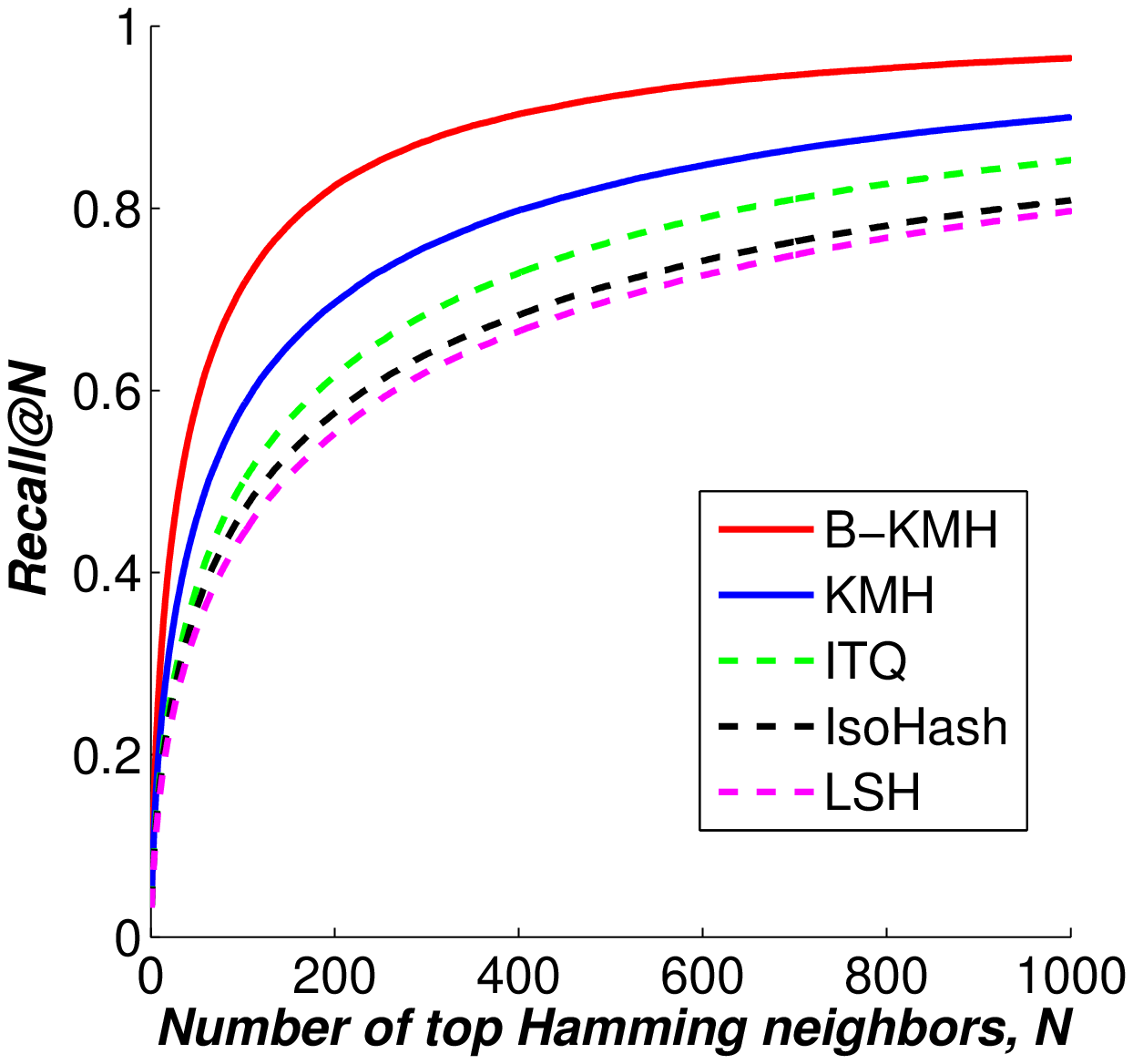} 
                \caption{\small{SIFT1M 128-bit.}}
                \label{fig:SIFT1M_128}
        \end{subfigure}%
        \\
        \begin{subfigure}[b]{0.445\textwidth}
                \centering
                \includegraphics[width=\textwidth]{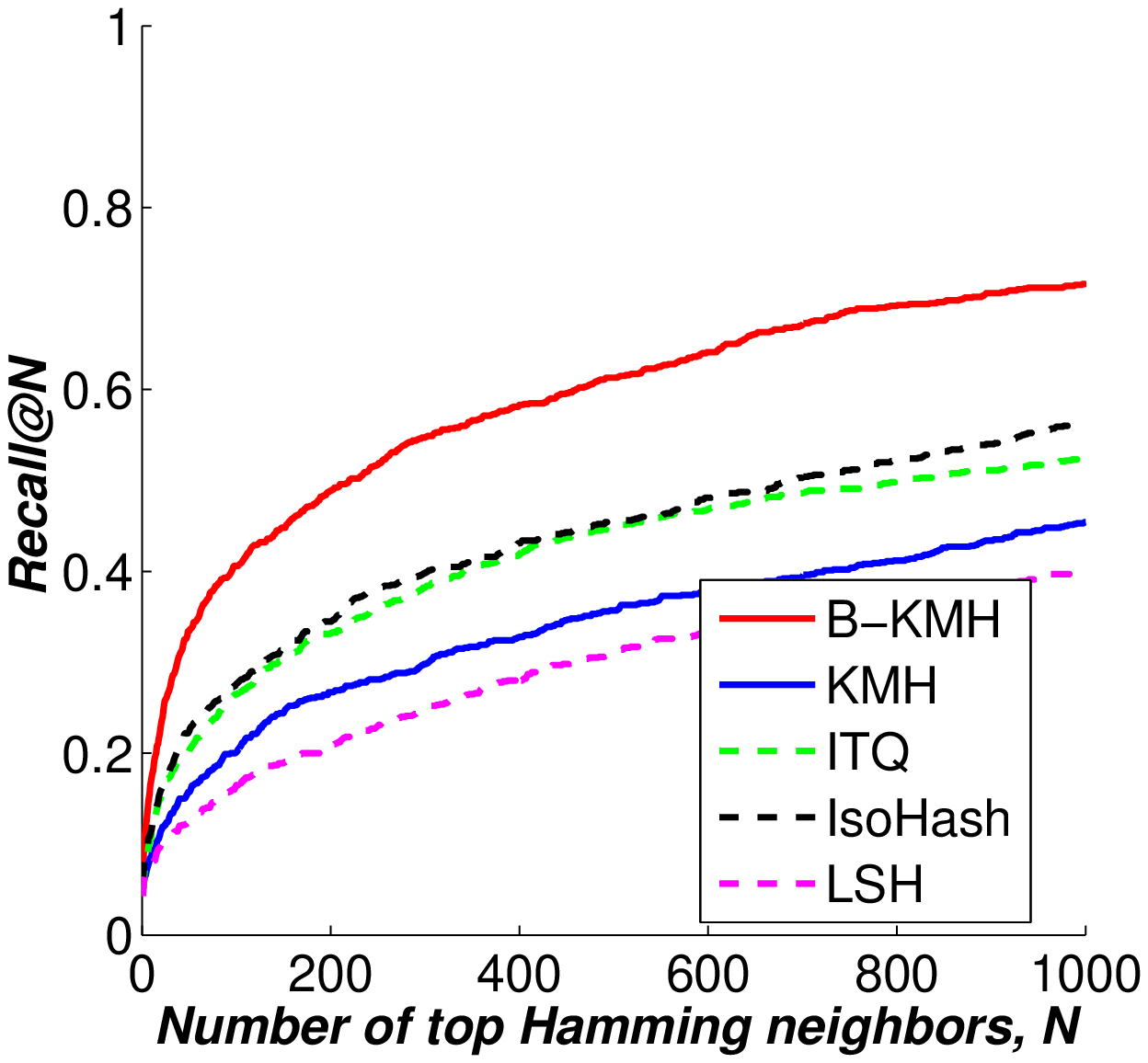}
               \caption{\small{GIST1M 64-bit.} }
               \label{fig:GIST1M_64}
        \end{subfigure} 

				\begin{subfigure}[b]{0.445\textwidth}
                \centering
                \includegraphics[width=\textwidth]{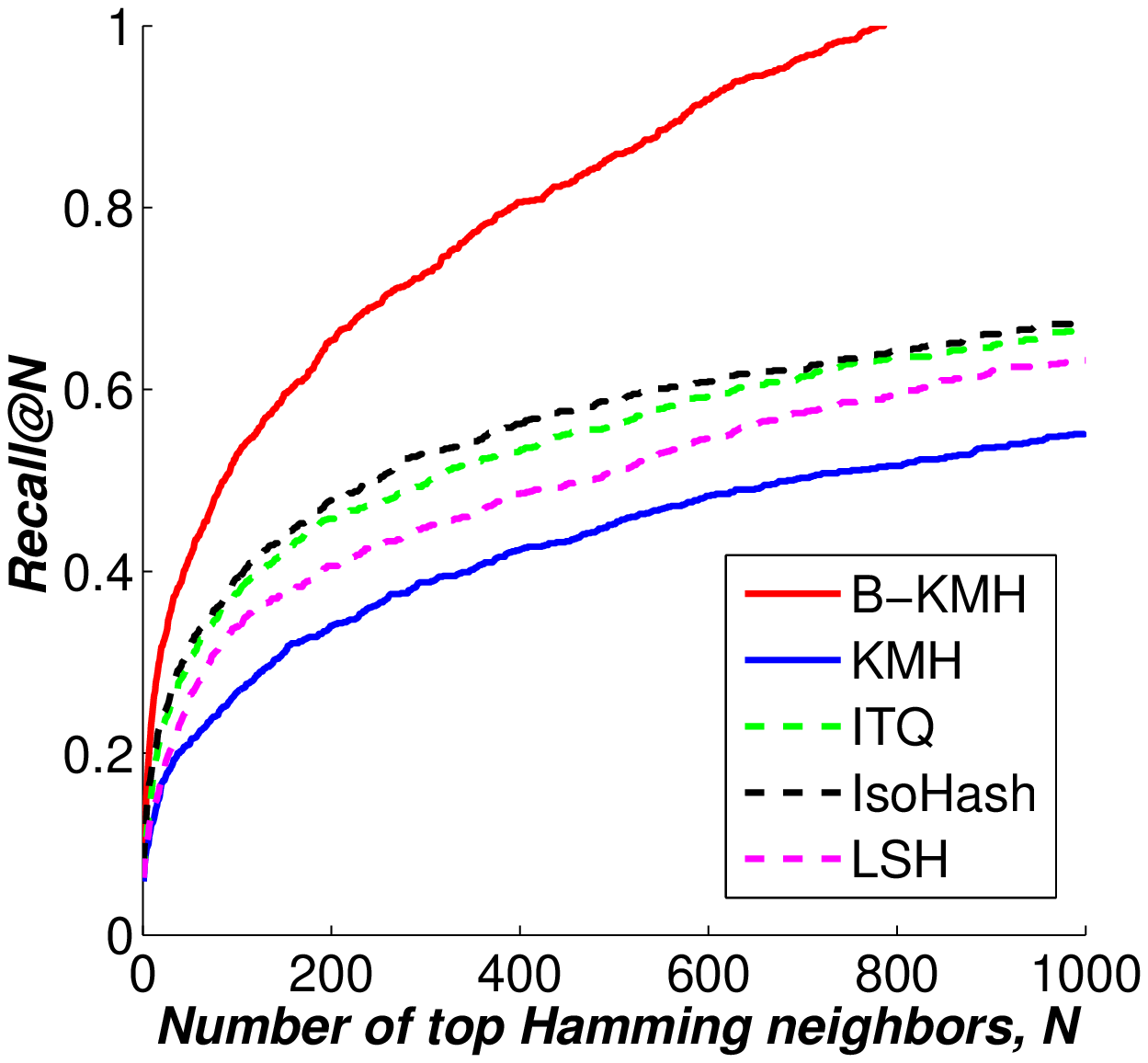} 
                \caption{\small{GIST1M 128-bit.}}
                \label{fig:GIST1M_128}
        \end{subfigure}%
        \end{array}$ 

        \caption{\small{Retrieval Performance on SIFT1M and GIST1M. $b=4$ and $\beta=8$ are used by KMH and B-KMH respectively for both 64-bit and 128-bit cases. We set $K=10$ here.}}
        \label{fig:SIFT1M_recall}
\end{figure*}

\section{Experiments}
We evaluate the ANN search performance on the SIFT1M and GIST1M datasets \cite{Jegou11}. Both datasets contain one million base points, and 10,000 and 1,000 queries respectively.  We consider the ground truth as each query's $K$ Euclidean nearest neighbors. We consider $K=1$, 10, and 100 in our experiments. 

We follow the search strategy of Hamming ranking commonly adopted in KMH and many hashing methods, where we sort the data according to their Hamming distances to the query. We evaluate the recall@$N$, where $N$ is the number of top Hamming neighbors. The recall is defined as the fraction of retrieved true nearest neighbors to the total number of true nearest neighbors $K$.

We compare the proposed B-KMH with KMH \cite{He13}, and some well-known hashing methods including iterative quantization (ITQ) \cite{Gong11}, isotropic hashing (IsoHash) \cite{Kong12}, and locality sensitive hashing (LSH) \cite{Andoni08}. All methods have publicly available codes and we use their default settings except KMH where we present the best test performance among multiple candidate values of $\lambda$.

In the SIFT1M experiments shown in Fig.~\ref{fig:SIFT1M_64} and \ref{fig:SIFT1M_128}, B-KMH outperforms KMH consistently in both 64-bit and 128-bit settings, which in turn outperforms other hashing algorithms also by a large margin. In particular, B-KMH improves the recall@1000 from 0.90 to 0.97. 

In the GIST1M experiments shown in Fig.~\ref{fig:GIST1M_64} and \ref{fig:GIST1M_128}, B-KMH significantly outperforms all other competing methods. For the 128-bit codes, B-KMH achieves perfect recall at $N=800$. Moreover, KMH performs much worse in the GIST1M dataset, presumably because adjusting codewords to respect the geometry of their binary representations causes $E_{\text{quan}}$ to be large.

Results in Fig.~\ref{fig:SIFT1M_recall} are generated using $b=4$. Therefore both KMH and B-KMH learn $k=16$ codewords in each subspace (16 subspaces for the 64-bit and 32 subspaces for 128-bit), but B-KMH uses $\beta=8$ bits to represent each codeword in computing the Hamming distance. The actual storage of the hash codes incur no addition cost because redundancy is removed by VLH. Therefore, the significant increase in performance is obtained with no increase in storage and marginal increase in computational cost.

\begin{figure*}[th] 
        \centering
				$
        \begin{array}{cc}

        \begin{subfigure}[b]{0.45\textwidth}
                \centering
                \includegraphics[width=\textwidth]{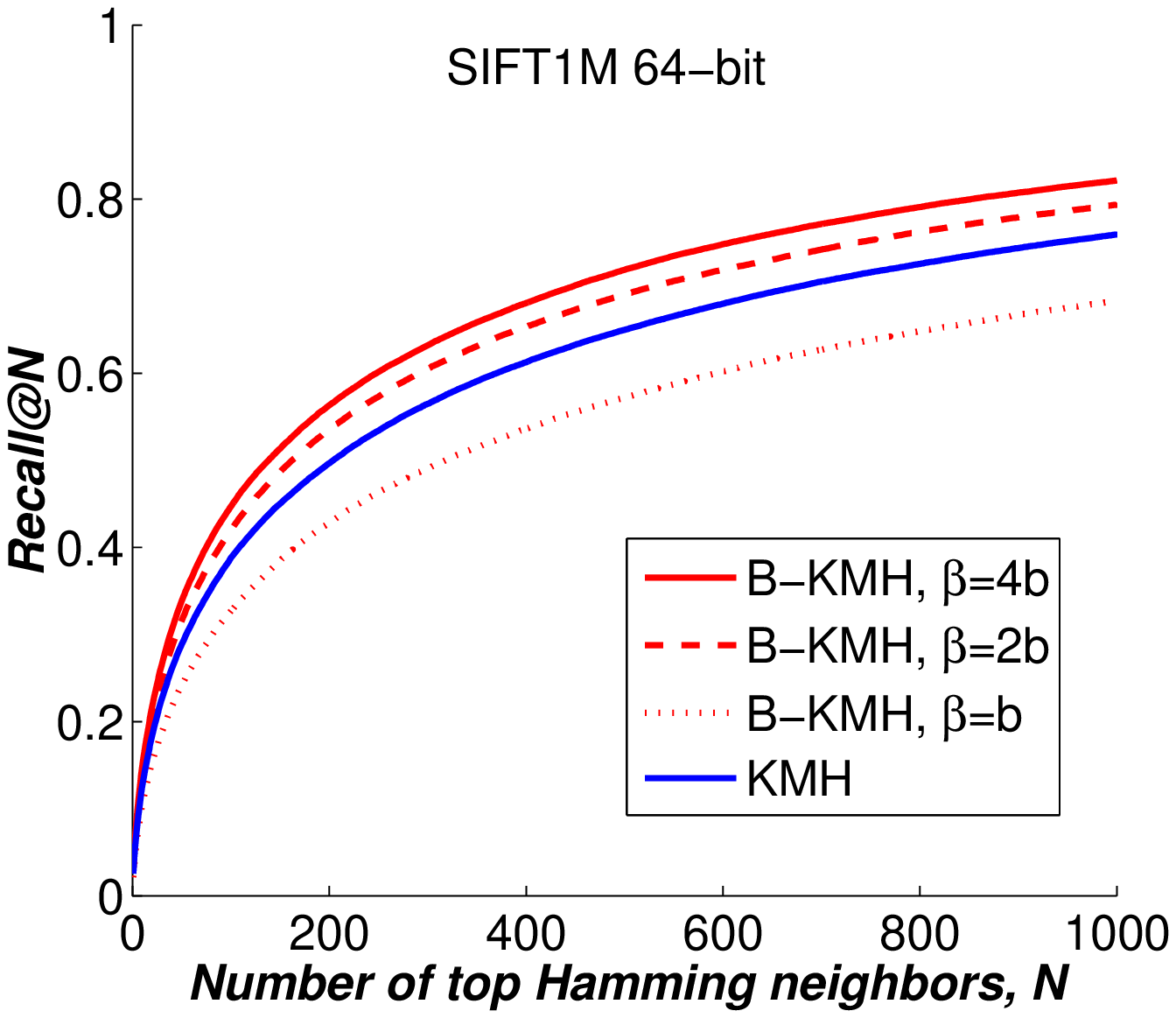}
                \caption{\small{Performance at different values of $\beta$ with $K=10$.}}
                \label{fig:SIFT1M_beta}
        \end{subfigure} 
        ~
        \begin{subfigure}[b]{0.45\textwidth}
                \centering
                \includegraphics[width=\textwidth]{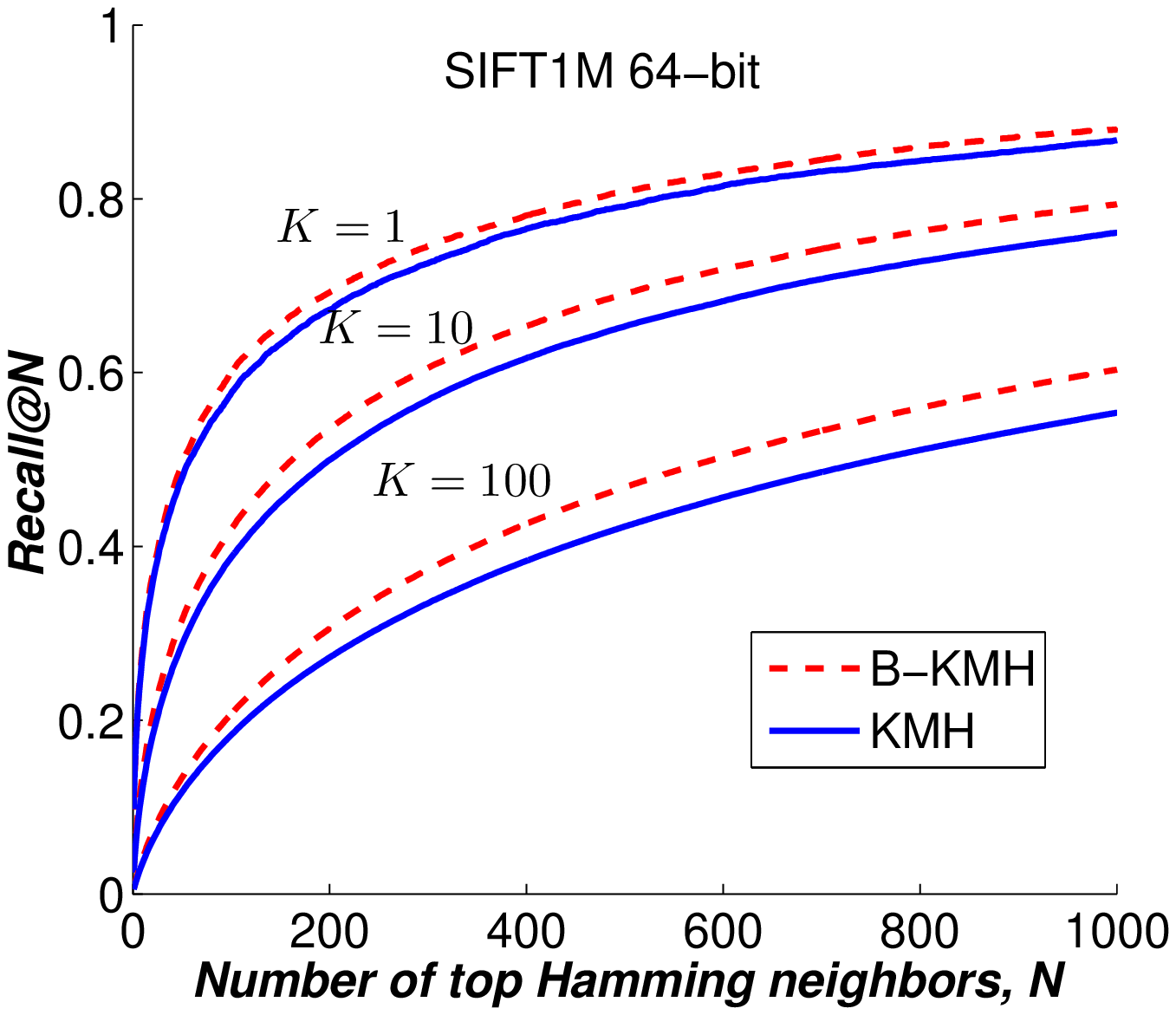}
                \caption{\small{Performance at different values of $K$ with $\beta=2b$.}}
                \label{fig:SIFT1M_K}
        \end{subfigure} 
        \end{array}$ 
        \caption{\small{Retrieval performance with different values of $\beta$ and $K$. Here $b=4$.}}
        \label{fig:SIFT1M_parameter}
\end{figure*}

In Fig.~\ref{fig:SIFT1M_beta}, we evaluate performance at various values of $\beta$. When $\beta=b$, B-KMH is similar to the naive two-step method, and the performance is worse than KMH. As we use more bits to represent codewords, B-KMH outperforms KMH, and we can observe an increased performance improvement with larger $\beta$. Though larger $\beta$ increases training and search complexity, the actual storage costs are the same for B-KMH and KMH. Figure \ref{fig:SIFT1M_K} compares recall performances at different ground truth nearest neighbor thresholds. It is clear B-KMH is superior across different $K$.

\section{Conclusion}
We have proposed a variable-length hashing (VLH) method which enables redundancy in hashing to be exploited in two ways. First, the hash can be compressed losslessly to reduce storage cost while incurring a marginal increase in search complexity. Second, redundancy can be deliberately introduced in the hash function design to improve retrieval performance without increasing storage cost. We have demonstrated the latter feature using K-means hashing, and believe that this strategy can be applied to other hash codes.

\newpage
\bibliographystyle{IEEEtran}
\bibliography{NIPS_ref}

\begin{thebibliography}{10}
\providecommand{\url}[1]{#1}
\csname url@samestyle\endcsname
\providecommand{\newblock}{\relax}
\providecommand{\bibinfo}[2]{#2}
\providecommand{\BIBentrySTDinterwordspacing}{\spaceskip=0pt\relax}
\providecommand{\BIBentryALTinterwordstretchfactor}{4}
\providecommand{\BIBentryALTinterwordspacing}{\spaceskip=\fontdimen2\font plus
\BIBentryALTinterwordstretchfactor\fontdimen3\font minus
  \fontdimen4\font\relax}
\providecommand{\BIBforeignlanguage}[2]{{%
\expandafter\ifx\csname l@#1\endcsname\relax
\typeout{** WARNING: IEEEtran.bst: No hyphenation pattern has been}%
\typeout{** loaded for the language `#1'. Using the pattern for}%
\typeout{** the default language instead.}%
\else
\language=\csname l@#1\endcsname
\fi
#2}}
\providecommand{\BIBdecl}{\relax}
\BIBdecl

\bibitem{Weiss08}
Y.~Weiss, A.~Torralba, and R.~Fergus, ``Spectral hashing,'' in \emph{NIPS},
  2008.

\bibitem{Indyk98}
P.~Indyk and R.~Motwani, ``Approximate nearest neighbors: Towards removing the
  curse of dimensionality,'' in \emph{ACM STOC}, 1998.

\bibitem{Shakhnarovich03}
G.~Shakhnarovich, P.~Viola, and T.~Darrell, ``Fast pose estimation with
  parameter sensitive hashing,'' in \emph{In ICCV}, 2003.

\bibitem{Raginsky09}
M.~Raginsky and S.~Lazebnik, ``Locality-sensitive binary codes from
  shift-invariant kernels,'' in \emph{NIPS}, 2009.

\bibitem{WangCVPR10}
J.~Wang, S.~Kumar, and S.-F. Chang, ``Semi-supervised hashing for scalable
  image retrieval,'' in \emph{CVPR}, 2010.

\bibitem{Liu12}
X.~Liu, J.~He, D.~Liu, and B.~Lang, ``Compact kernel hashing with multiple
  features,'' in \emph{ACM MM}, 2012.

\bibitem{Liu13}
X.~Liu, J.~He, B.~Lang, and S.-F. Chang, ``Hash bit selection: {A} unified
  solution for selection problems in hashing,'' in \emph{CVPR}, 2013.

\bibitem{YuICASSP15}
H.~Yu and P.~Moulin, ``{SNR} maximization hashing for learning compact binary
  codes,'' in \emph{ICASSP}, 2015.

\bibitem{Wang10}
J.~Wang, S.~Kumar, and S.-F. Chang, ``Sequential projection learning for
  hashing with compact codes,'' in \emph{ICML}, 2010.

\bibitem{Gong11}
Y.~Gong and S.~Lazebnik, ``Iterative quantization: A procrustean approach to
  learning binary codes,'' in \emph{CVPR}, 2011.

\bibitem{liu14}
W.~Liu, C.~Mu, S.~Kumar, and S.-F. Chang, ``Discrete graph hashing,'' in
  \emph{NIPS}, 2014.

\bibitem{Yu15}
H.~Yu and P.~Moulin, ``{SNR} maximization hashing,'' \emph{IEEE TIFS}, 2015.

\bibitem{Shakhnarovich05}
G.~Shakhnarovich, ``Learning task-specificsimilarity,'' \emph{PhD dissertation,
  MIT}, 2005.

\bibitem{Lin10}
R.-S. Lin, D.~A. Ross, and J.~Yagnik, ``{SPEC} hashing: Similarity preserving
  algorithm for entropy-based coding,'' in \emph{CVPR}, 2010.

\bibitem{Kong12}
W.~Kong and W.-J. Li, ``Isotropic hashing,'' in \emph{NIPS}, 2012.

\bibitem{Liu11}
W.~Liu, J.~Wang, and S.~fu~Chang, ``Hashing with graphs,'' in \emph{In ICML},
  2011.

\bibitem{KulisNIPS09}
B.~Kulis and T.~Darrell, ``Learning to hash with binary reconstructive
  embeddings,'' in \emph{NIPS}, 2009.

\bibitem{Weiss12}
Y.~Weiss, R.~Fergus, and A.~Torralba, ``Multidimensional spectral hashing,'' in
  \emph{ECCV}, 2012.

\bibitem{Norouzi12}
M.~Norouzi, D.~J. Fleet, and R.~Salakhutdinov, ``Hamming distance metric
  learning,'' in \emph{NIPS}, 2012.

\bibitem{Cover:1991}
T.~M. Cover and J.~A. Thomas, \emph{Elements of Information Theory, 2nd
  Edition}.\hskip 1em plus 0.5em minus 0.4em\relax Wiley-Interscience, 2006.

\bibitem{Norouzi12MultiIndex}
M.~Norouzi, A.~Punjani, and D.~J. Fleet, ``Fast search in {Hamming} space with
  multi-index hashing,'' in \emph{CVPR}, 2012.

\bibitem{Aharon06}
M.~Aharon, M.~Elad, and A.~Bruckstein, ``{K-SVD}: An algorithm for designing
  overcomplete dictionaries for sparse representation,'' \emph{IEEE TSP}, 2006.

\bibitem{Haitsma02}
J.~Haitsma and T.~Kalker, ``A highly robust audio fingerprinting system,'' in
  \emph{ISMIR}, 2002.

\bibitem{He13}
K.~He, F.~Wen, and J.~Sun, ``K-means hashing: An affinity-preserving
  quantization method for learning binary compact codes,'' in \emph{CVPR},
  2013.

\bibitem{Sabin84}
M.~Sabin and R.~Gray, ``Product code vector quantizers for waveform and voice
  coding,'' \emph{IEEE TASSP}, 1984.

\bibitem{Jegou11}
H.~J\'egou, M.~Douze, and C.~Schmid, ``Product quantization for nearest
  neighbor search,'' \emph{IEEE TPAMI}, 2011.

\bibitem{Lowe04}
D.~G. Lowe, ``Distinctive image features from scale-invariant keypoints,''
  \emph{IJCV}, 2004.

\bibitem{Andoni08}
A.~Andoni and P.~Indyk, ``Near-optimal hashing algorithms for approximate
  nearest neighbor in high dimensions,'' \emph{Commun. ACM}, 2008.

\end{thebibliography}

\end{document}